\documentclass[conference]{IEEEtran}
\IEEEoverridecommandlockouts
\usepackage{cite}
\usepackage{amsmath,amssymb,amsfonts}
\usepackage{algorithmic}
\usepackage{graphicx}
\usepackage{textcomp}
\usepackage{subfig}
\usepackage{tikz}
\usepackage{booktabs}
\usepackage[normalem]{ulem}
\useunder{\uline}{\ul}{}
\usepackage{lipsum,adjustbox}
\usepackage{multirow}
\usepackage{balance}
\usepackage{comment}

\usepackage{float} 
\usepackage{caption}

\usepackage{xcolor}
\def\BibTeX{{\rm B\kern-.05em{\sc i\kern-.025em b}\kern-.08em
    T\kern-.1667em\lower.7ex\hbox{E}\kern-.125emX}}
    
\usepackage{fancyhdr}
\begin{document}

\title{End-to-End Speech Emotion Recognition: Challenges of Real-Life Emergency Call Centers Data Recordings}


\author{\IEEEauthorblockN{Théo Deschamps-Berger }
\IEEEauthorblockA{\textit{LISN} \\
\textit{Paris-Saclay University, CNRS}\\
Orsay, France \\
theo.deschamps-berger@u-psud.fr}
\and
\IEEEauthorblockN{Lori Lamel }
\IEEEauthorblockA{\textit{LISN} \\
\textit{CNRS}\\
Orsay, France \\
lori.lamel@limsi.fr}
\and
\IEEEauthorblockN{Laurence Devillers }
\IEEEauthorblockA{\textit{LISN} \\
\textit{CNRS}\\
Orsay, France \\
devil@limsi.fr}

}

\maketitle
\thispagestyle{fancy}

\begin{abstract}

Recognizing a speaker’s emotion from their speech can be a key element in emergency call centers. End-to-end deep learning systems for speech emotion recognition now achieve equivalent or even better results than conventional machine learning approaches. 
In this paper, in order to validate the performance of our neural network architecture \textbf{for emotion recognition from speech}, we first trained and tested it on the widely used corpus accessible by the community, IEMOCAP. We then used the same architecture with the real life corpus, CEMO, comprised of 440 dialogs (2h16m)  from 485 speakers. The most frequent emotions expressed by callers in these real-life emergency dialogues are fear, anger and positive emotions such as relief. In the IEMOCAP general topic conversations, the most frequent emotions are sadness, anger and happiness. Using the same end-to-end deep learning architecture, an Unweighted Accuracy Recall (UA) of 63\% is obtained on IEMOCAP  and a UA of 45.6\%  on CEMO, each with 4 classes. Using only 2 classes (Anger, Neutral), the results for CEMO are 76.9\% UA compared to 81.1\% UA for IEMOCAP. We expect that these encouraging results with CEMO can be improved by combining the audio channel with the linguistic channel. Real-life emotions are clearly more complex than acted ones, mainly due  to  the large diversity of emotional expressions of speakers.
\end{abstract}

\begin{IEEEkeywords}
emotion detection, end-to-end deep learning architecture, call center, real-life database, complex emotions.
\end{IEEEkeywords}

\section{Introduction}
Detecting the speaker’s emotion can be a key element in many applications, notably in emergency call centers. Very few studies have addressed the detection of natural emotions in real-world conversations. For example, the Audiovisual Interest Corpus (AVIC) \cite{b1}, the reality TV \cite{b2} or the SEWA DB \cite{b18} are considered as naturalistic data. However, most current emotion research is still conducted on artificial corpora with intentionally balanced emotions that are collected in laboratory or simulated settings, and include speech from only a small number of speakers, e.g. IEMOCAP \cite{b3} or MSPImprov \cite{b4}. 

In this paper a state-of-the-art deep learning system is tested on a large real-life database of  calls in French to an medical emergency center CEMO \cite{b5}. 
Due to the number of speakers and the natural context of the collection, a large amount of variability exists in the dialogs comprising this corpus. Sometimes there are more than one caller per dialog (e.g. a family member of the caller), with a lot of blended emotions and shaded feelings. The quality of the recording is often quite poor, the amount of emotional data quite low, and usually there are only a few words spoken by each speaker.

Our aim is the detection of emotions in real-life speech for use in a real application, that is, an emergency call center \cite{b19}, \cite{b20}. The envisaged usage is to enrich the dashboard of the agents with on-line speaker's emotional state detection from callers, to help them with decision making. In contrast to most recent published studies \cite{b6}, \cite{b7} conducted on corpora with few speakers such as IEMOCAP or MSP-Improv, this paper addresses the challenge of real-life emotions with a large set of speakers. In order to be comparable with results obtained in the community, we first tested  with the IEMOCAP corpus to optimize a deep learning architecture for speech emotion recognition (SER). Then we used the same architecture with the CEMO corpus \cite{b5}, \cite{b8}. 

Early systems for emotion detection were often built using open source tools for acoustic feature extraction and a classical approach such as SVM classifiers. More recently, many state-of-the-art AI systems for emotional detection use an end-to-end deep learning architecture combining audio and linguistic cues \cite{b6}, \cite{b7}. In this paper, we focus on the emotion detection task in speech, without explicit linguistic information. Usually, Convolutional Neural Networks or Recurrent Neural Networks are used to detect near and long dependencies in utterances \cite{b9}. The system can also be combined with highway connectivity to handle noisy conversations via discriminative learning of the representation \cite{b10}. Several other optimizations have been proposed, mostly tested on the widely used  IEMOCAP database: multitask learning, concatenation of $\Delta\Delta_2$ to the spectrograms \cite{b11} or attention mechanisms: either self-attention mechanism \cite{b12} or Multi-head attention mechanism \cite{b13} to sort out the salience of each part of the sentence. 

Inspired by the recent achievements in speech emotion detection with end-to-end approaches \cite{b9}, \cite{b11}, a mixed Convolutional Neural Network and Bidirectional Long Short Term Memory (CNN-BiLSTM) architecture is explored in this work. The main originality of our paper is training and testing the same end-to-end architecture that achieves competitive results on the widely used IEMOCAP (Spontaneous portion) corpus, on the realistic CEMO data. The two databases are presented in Section 2, followed by a description of the selected deep learning architecture in Section 3. Section 4 overviews the experimental conditions and presents results, followed by conclusions and directions for future research in Section 5.

\section{Databases}

Although the main aim of this study is speech emotion detection for an emergency call center application,
in order to compare our results with other published research, the same end-to-end system was explored with the 2 databases described in this section: one is the spontaneous portion of the well known IEMOCAP database, the other a  real-life database CEMO from the targeted task.

\subsection{IEMOCAP}

The Interactive Emotional Dyadic Motion Capture (IEMOCAP), collected at the University of Southern California (USC) \cite{b3}, one of the standard databases for emotion studies, was used to test the end-to-end architecture. It consists of twelve hours of audio-video recordings performed by 10 professional actors (five women and five men) and organized in 5 sessions of dialogues between two actors of different genders, either acting out a script or improvising. Each sample of the audio set is an utterance with an associated emotion label. Labeling was made by three USC students for each utterance. The annotators were allowed to assign multiple labels if necessary. The final 'true' label for each utterance was chosen by a majority vote if the emotion category with the highest vote was unique. Since the annotators reached consensus more often when labeling the improvised utterances (83.1\%) than the scripted ones (66.9\%) \cite{b3}, \cite{b17}, we only used the improvised part of the speech database.
For comparison with previous state-of-the-art approaches, four of the most represented emotions: neutral, sadness, anger and happiness are predicted, leaving us with 2280 utterances in total (2h48mn). The average audio segment is 4.4s (median 3.5s, min=0.7s, max=29.1s).

\subsection{CEMO}

Call center data is a particular form of natural data collected in a real-life context. The recording is imperceptible to the speakers and therefore does not affect the spontaneity of the data. Moreover, with telephone data, emotion expression can only be assessed via the voice with no possibility of support or conflict from other modalities such as actions, gestures or facial expressions which are available in the IEMOCAP videos. The CEMO corpus contains 20 hours of recordings of real conversations between agents and callers \cite{b5}, \cite{b8}. 
The service, whose role is to give medical advice, can be contacted 24 hours a day, 7 days a week.  During an interaction, an agent will use a precise and predefined strategy to obtain information in the most efficient way possible. The agent's role is to determine the subject of the call and to quickly assess the its urgency, making an informed decision as to what action is required. The decision taken may be to send an ambulance, to redirect the caller to social or psychiatric center, or to advise the caller to take a followup action, e.g. to go to the hospital or to call their doctor. The caller may be the patient or a third party (family, friend, colleague, neighbor). In the case of urgent calls, the caller will often express stress, pain, fear, or even panic but may also express annoyance or even anger towards the medical regulatory agents during the call. A list of 21 fine-grained labels was used  to provide annotations at a segment level which is often smaller than speaker turn. The fine labels were also merged into 7 coarse-grained emotion labels (macroclasses): Fear (Fear, Anxiety, Stress, Panic, Embarrassment, Dismay), Anger (Annoyance, Impatience, HotAnger, ColdAnger), Sadness (Disappointment, Sadness, Despair, Resignation), Pain, Positive (Interest, Compassion, Amusement, Relief), Surprise and Neutral. During the annotation phase, the coders were given the possibility to choose two labels in order to describe complex emotions.  Only about 30\% of the segments were annotated with an emotion label (from agents and callers).
In order to assess the consistency of the selected labels, the inter-annotator agreement (between 2 coders) has been calculated. The Kappa value is 0.61 for callers and 0.35 for agents when only considering the Major macro-classes annotation. The Kappa values are slightly better (0.65 and 0.37, respectively) if the following rule is used: it is necessary to have at least one common label between the annotations of the two coders (Major or Minor). 
The annotation is seen to be much more reliable for the caller's speech than for that of the agent, which may be due to their respective goals and roles: the callers contact the medical service for a specific task (get help or information), and the Agent, in the context of his/her job, has to control the dialog so as to obtain the required information about the caller and help him/her.

\begin{figure}[htbp]%
    \centering
    \subfloat[\centering IEMOCAP (2280  segments)]{{\includegraphics[width=3.8cm]{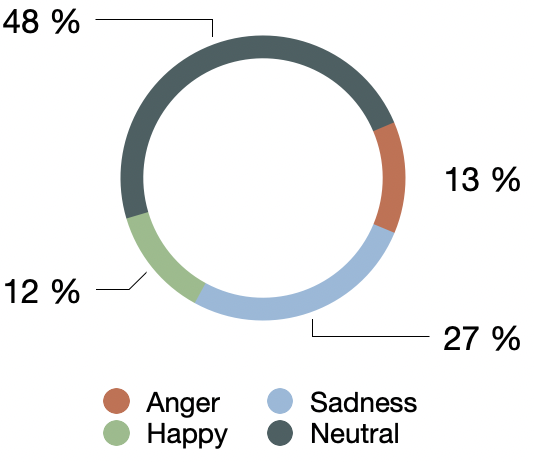} }}%
    \qquad
    \subfloat[\centering CEMO-4eC (6931 segments)]{{\includegraphics[width=3.8cm]{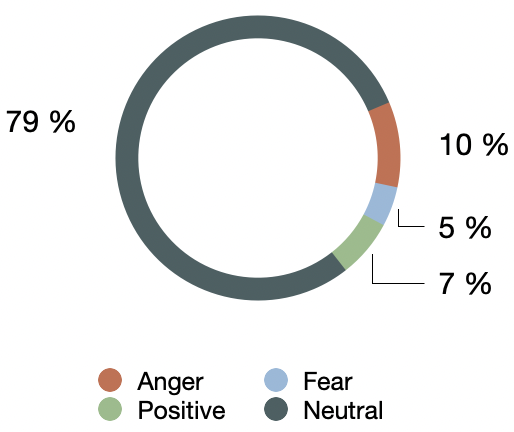} }}%
    \caption{Distribution of the 4 emotions in IEMOCAP and in CEMO-4eC} %
    \label{1}%
\end{figure}

The 4 most frequent coarse-grained emotion labels were used for CEMO: Neutral, Anger, Positive and Fear. After restricting the CEMO data to these 4 emotions, we obtained a subset of 6931 segments from 807 callers, excluding turns of the agents as they rarely exhibit emotions (as required by their role).
The distribution of the 4 emotion labels is this data subset is shown in Fig.~\ref{1}. It can be seen that there is a large class imbalance in the CEMO data, with almost 80\% of the segments labeled as neutral.

To reduce the large class imbalance, callers for whom all segments were labeled as neutral were excluded from this study as described in the next section. The resulting subset of the corpus contains 440 dialogues from 485 callers (159 male, 326 female) (2h16mn), with a total of 4825 segments from callers with the macro-emotions: Fear, Anger, Positive and Neutral. The average audio segment duration is 1.7s (median 1.1s, min=0.3s, max=22.8s). 

\subsection{Comparing a corpus created for research and a real-life corpus}

Based on the descriptions above, there are several notable differences between a corpus created for research purposes (IEMOCAP) and a corpus collected in an emergency call center (CEMO). These differences concern the number of speakers and their characteristics (gender, age, relationship with patient), the amount of speech for each and the distribution of emotions.

In IEMOCAP we used the 4 most frequent emotions (2280 segments) of the spontaneous part, as shown in Fig.~\ref{1}(a). 
For the CEMO corpus, we selected the 6931 segments from callers annotated with one of the four emotions (we refer to this as CEMO-4eC: CEMO-4emotions from Callers). There are only 22\% of non-neutral segments as can be seen in part (b) of Fig.~\ref{1}. For CEMO-4eC, the average number of segments per caller is 13 (the median is 12 segments (min=1, max=46 segments), whereas for IEMOCAP, there are more segments per speaker, with an average number of 236 segments per speaker (the median is 221). Furthermore, the average audio segment duration is shorter for CEMO-4eC (1.7s) than for IEMOCAP (4.4s).

\begin{figure}[htbp]
    \centering
    \includegraphics[width=0.4\textwidth]{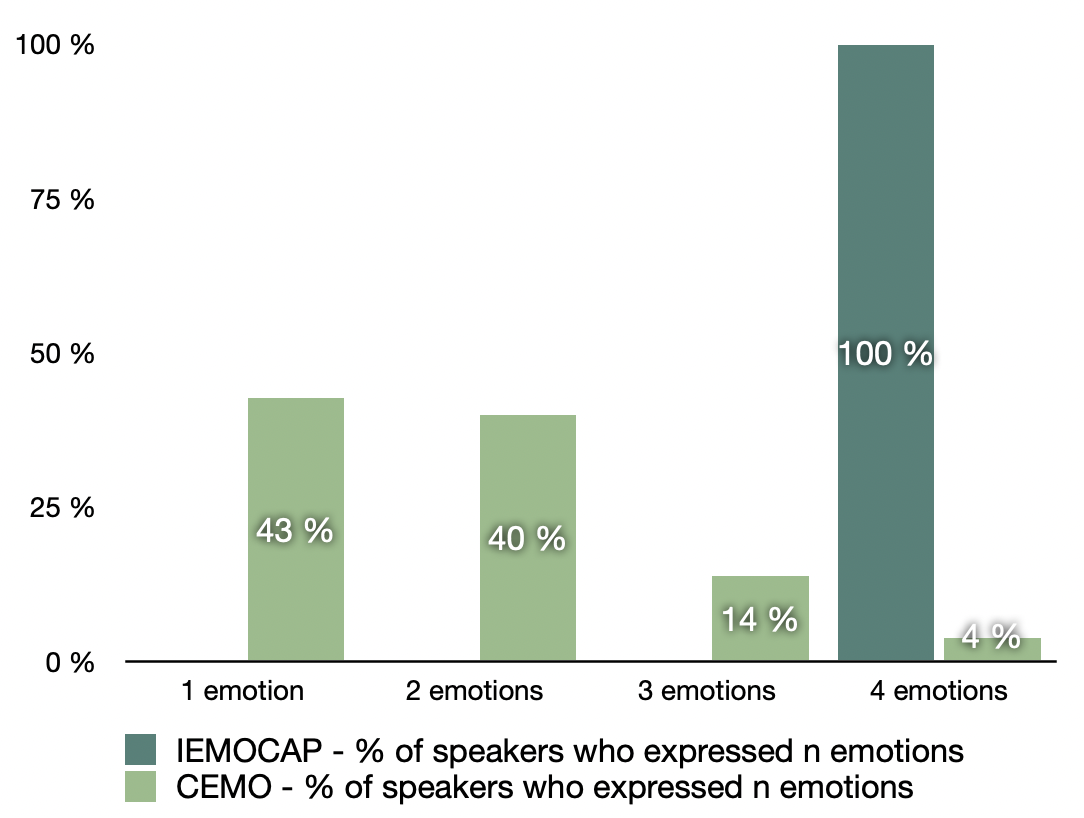}
    \caption{Percentage of speakers expressing 4, 3, 2 or only 1 emotion in IEMOCAP (10 speakers) and CEMO-4eC (807 speakers)}
    \label{2}
\end{figure}

As can be seen in Fig.~ \ref{2}, in the CEMO-4eC corpus, only 4\% of speakers expressed the 4 emotions instead of all of them in IEMOCAP. 
The  (about 40\%) show either 1 or 2 emotions. The distribution by gender is also different: 50\% men, 50\% women for IEMOCAP versus 35\% men and 65\% women for CEMO-4eC (the caller distribution in the full CEMO corpus is similar).

\subsection{Balanced CEMO-4eC Corpus for training}
\begin{figure}[htbp]%
    \centering
    \centering  \includegraphics[width=9.cm]{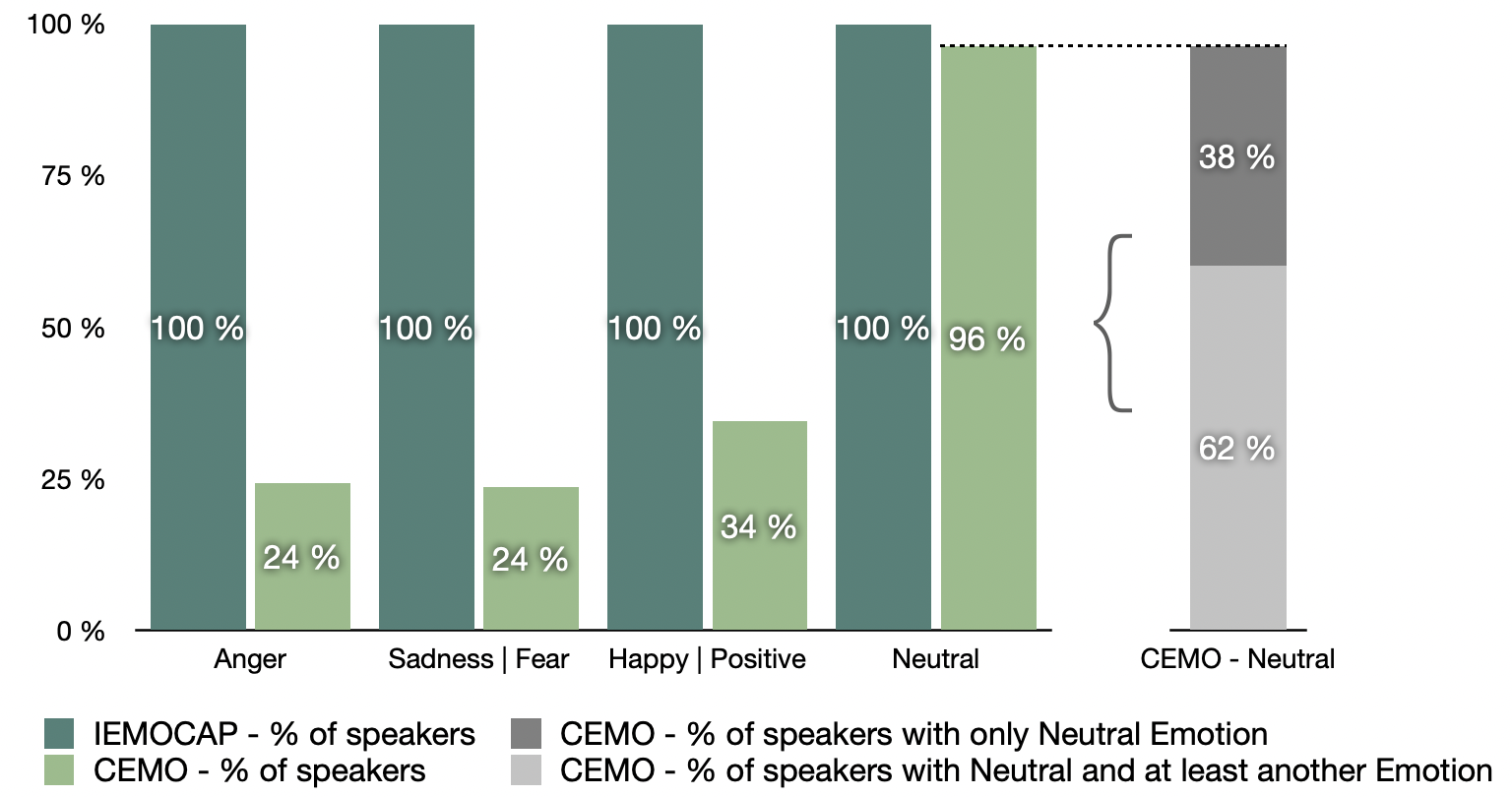} %
    \caption{\textit{\textbf{Left}}: Percentage of speakers expressing a specific emotion class in both databases, IEMOCAP (10 speakers) and CEMO-4eC (807 speakers). \textit{\textbf{Right}}: Percentage of speakers in the CEMO neutral class (778 speakers) who have only neutral segments vs at least one emotional segment.}
    \label{3}%
\end{figure}

Looking more closely at the two corpora, Fig.~\ref{3} shows the percentage of speakers expressing a specific emotion class in both databases.
Indeed, all speakers in IEMOCAP have utterances covering the 4 different emotions, whereas only a minority of the speakers in CEMO-4eC expressed any emotion.

As mentioned earlier, to mitigate the problem of imbalance in CEMO-4eC, 38\% of speakers (322 speakers) of the neutral class (Fig. \ref{3})) who were judged by the annotators to have produced only neutral segments were excluded for the remainder of our studies. The resulting subset of the CEMO-4eC database contains the speaker turns from the 485 callers from 440 dialogues which we will refer to as CEMO-4eC$_s$ in the remainder of this paper.

\section{End-to-end deep learning architecture}
\begin{figure*}[htbp]
    \centering
    \includegraphics[width=11.4cm,height=13cm]{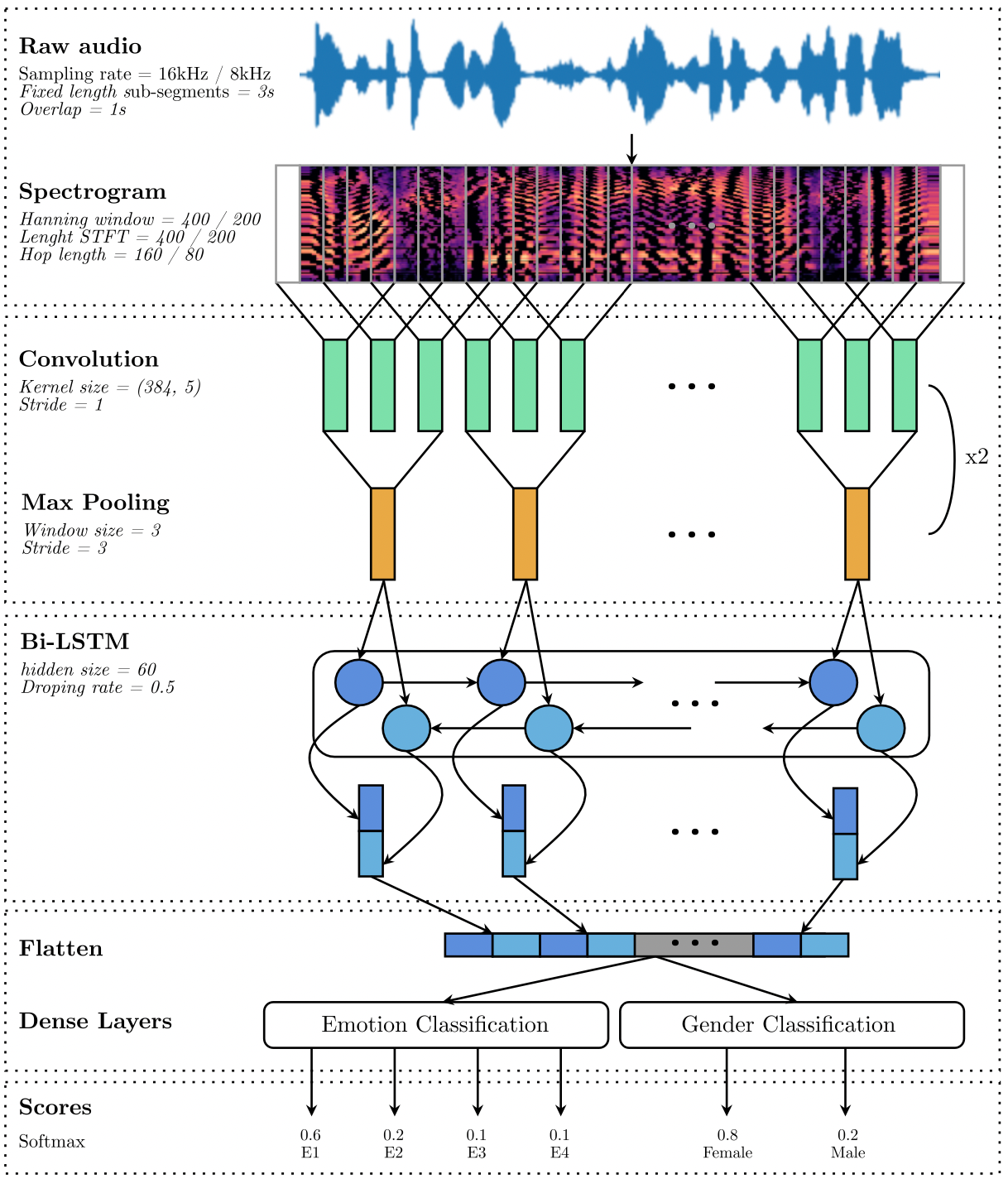}
    \caption{End-to-end Temporal CNN-BiLSTM}
    \label{4}
\end{figure*}

This section describes the deep learning architecture chosed for this study.
We constructed an end-to-end CNN-BiLSTM system to predict emotions from the raw audio signals using the architecture shown in Fig.~\ref{4}.

\subsection{Preprocessing}

Preprocessing has two main steps, feature extraction followed by chopping and sampling the segments.

\subsubsection{Spectral feature extraction}

For each audio signal (sampling rate: 16kHz for IEMOCAP, 8kHz for CEMO-4eC$_s$), a Hanning window of length of 25ms is applied. Then, a Short Term Fourier Transform (STFT) of length 10ms offset is computed. The STFT is then mapped to Mel's scale. Finally, the $\Delta\Delta_2$ of the STFT are concatenated as input to the system.
Computing first and second order Delta parameters is a common method to determine the changes of the spectral features over time.

\subsubsection{Sub-segment sampling}

For both corpora, each audio segment was split in sub-segments of 3s as proposed by \cite{b9}. Since there can be a large variation in how much of the full audio segment expresses emotion, cues may be found in only one or in several sub-segments. Therefore, some classes of emotions such as Anger or Fear for example in CEMO-4eC$_s$ will be present in more sub-segments than segments. 

The sampling method was extended by using an overlap of 1s in order to avoid cutting contextual emotion information. Tests using segment sizes ranging between 1s and 4s did not lead to an improvement on either corpus, so we decided to use 3s for both systems. The last sub-segments under 3s are padding with zeros to maintain a fixed length. This fixed length for each input is necessary to perform the convolutions for our architecture. The final distribution used for training is given in Table~\ref{tab:table1}. We assigned the label of a segment to all of the created sub-segments. The speech segments in the two corpora have different lengths: 4.4s in average on IEMOCAP, 1.7s for CEMO-4eC$_s$. In order to have about the same number of segments in each class, we decreased the size of the neutral class, being careful to keep at least one neutral sample for each caller. %
Then oversampling was used for the training phase in order to have equal number of segments per class.

\begin{table}[htbp]
\centering
\caption{Final distribution of segments/sub-segments used in the training phase for both corpora: IEMOCAP and CEMO-4eC$_s$.}
\label{tab:table1}
\begin{tabular}{@{}llll@{}}
\toprule
\textbf{IEMOCAP} & \textbf{\#seg./\#sub-seg.} & \textbf{CEMO-4eC$_s$} & \textbf{\#seg./\#sub-seg.} \\ \midrule
Anger            & 289 / 925         & Anger         & 672 / 1325        \\
Sadness          & 608 / 2176        & Fear          & 312 / 826         \\
Happy            & 284 / 822         & Positive      & 459 / 594         \\
Neutral          & 1099 / 3107        & Neutral       & 3382 / 3916        \\ \bottomrule
\end{tabular}
\end{table}

\subsection{CNN: Temporal or 2D convolution}

Convolution Neural Networks (CNN) are a reference in image classification. The intuition here is to consider the segments from the audio as images. The CNN layer identifies local contexts by applying  $n$ convolutions over the input audio images along the time axis and produces a sequence of vectors.
We explored two convolution kernels:
\begin{itemize}
\item a 2D CNN-BiLSTM  with 1,247,374 trainable parameters (for 4 classes) commonly used for vision. 
\item a Temporal CNN-BiLSTM as shown in Fig.~\ref{4} with 219,062 trainable parameters (for 4 classes) to take advantage of the temporal information, i.e. a specific kernel to perform a convolution along the time axis.

The Temporal CNN-BiLSTM was adopted for the rest of the paper due to its efficiency and slightly better results than the conventional 2D CNN in preliminary experiments (see Table.\ref{tab:table2}).
\end{itemize}

\subsection{BiLSTM with mask}

The attention mask aims to help the Bidirectional Long Short-Term Memory (BiLSTM) ignore information coming from the zero padding part convoluted in the CNN layers. The original size of the segment before padding is kept in memory to calculate the mask size. The mask size at the output of a convolution is calculated using these equations (H: Height, W: Width):%
\begin{align}
    Output_H = \frac{Input_{H}–Kernel_{W}+2*Padding}{Stride_{W}} + 1 \label{eq:3} \\*[3ex]
    Output_W = \frac{Input_{W}–Kernel_{H}+2*Padding}{Stride_{H}} + 1 \label{eq:4}
\end{align}
\\
The LSTM has the ability to weigh the information it receives and transmit it through gates. It is useful to locate long-term dependencies. We concatenate the output of the two LSTMs (computed from left to right and right to left), with 60 hidden units and a dropout of 50\% for the last one. We used the output of all the LSTM hidden cells as input to a Dense network. The dense network will learn which part of the segment best predicts the emotion.

\subsection{Multitask classification}

In addition to the emotion classification task, we incorporated contextual information to help predict emotions. The Temporal CNN-BiLSTM is tested on the one hand with emotion classification alone and on the other hand with a shared loss between emotion and gender which was reported to improve performance. \cite{b11}. The model is optimized by the following objective function: 
\begin{align}
   \mathcal{L}oss = \mathcal{L}oss_{emotion} + \mathcal{L}oss_{gender} \label{eq:5}
\end{align}

\subsection{Evaluation methodology}

Both systems use a 5-fold cross validation strategy, independent of the speaker. This means that, for example in IEMOCAP,  4 sessions are dedicated for training (8 speakers) and the last session is split for validation (1 speaker) and test (1 speaker). The same strategy is applied to CEMO-4eC$_s$ with more speakers.
During each fold, system training is optimized on the best Unweighted Accuracy Recall of the validation set.
During the testing phase we evaluate the prediction for the full segment by computing a Majority vote on each of the sub-segment predictions, and also by computing the average and maximum of the posterior probabilities of the respective sub-segments of one audio signal.
Depending on the predictions, we adopt the best strategy between majority voting, mean and max. 
The following measures are used for evaluation:  UA (Unweighted Accuracy Recall)  and WA (Weighted Accuracy Recall) (eqn. \ref{eq:6}-\ref{eq:8}).

\begin{align}\label{eq:6}
    Recall_i = \frac{TP_i}{TP_i+FN_i}\\*[3ex] 
    \label{eq:7}
    UA = \frac{\sum_{i=1}^E Recall_i}{E}\\*[3ex] 
    \label{eq:8}
    WA = \sum_{i=1}^E \frac{\#Samples_i}{N}*Recall_i 
\end{align}

\begin{itemize}
        \item $TP_i$ and $FN_i$ are the number of true positive and false negative instances respectively for emotion $i$
        \item $N$ is the total number of instances from all emotions
        \item $E$ is the total number of emotions
\end{itemize}

\section{Experiments and Results}

This section reports on and discusses the experimental results assessing the performance of our DNN systems for speech emotion recognition on the two databases, using 4 and then fewer emotion classes.

\subsection{IEMOCAP: Emotion detection on 4 classes}

We first verified the performance of our DNN on the spontaneous part of the widely used IEMOCAP corpus.
Table~\ref{tab:table2} shows the results obtained with the Temporal CNN-BiLSTM (Fig.~\ref{4}) and a CNN-BiLSTM. Our results are comparable to the performance obtained on the same database with 5-folds by \cite{b7} with CNN-BiLSTM and \cite{b14} with CNN-BiLSTM. Our best results are seen to be obtained with the Temporal CNN-BiLSTM. 

\begin{table}
\centering
\caption{5-fold cross validation scores with 4 emotions on the IEMOCAP improvised subset comparing the 2D CNN-BiLSTM and Temporal CNN-BiLSTM. For each experiment, the results corresponding to its best run are given. The top part of the table reports state-of-the-art published results, and the bottom our experiments.}
\label{tab:table2}
\begin{tabular}{@{}cc|c|c|c@{}}
\toprule
\textbf{}             &                       &       & \multicolumn{2}{c}{\textbf{IEMOCAP}} \\
\multicolumn{2}{c|}{\textbf{Cond. (4 emotions)}}                             & \textbf{\#par.} & \multicolumn{2}{c}{(Eng-US)} \\
                      &                       &       & \sc{ua} (\% )           & \sc{wa} (\% )            \\ \midrule
\multicolumn{1}{c|}{} & AE-BLSTM {\cite{b15}} & --    & 52.8          & 54.6         \\
\multicolumn{1}{c|}{State-of-the-art}              & CNN-biLSTM {\cite{b7}} & --             & 59.4          & 68.8          \\
\multicolumn{1}{c|}{} & RNN-ELM {\cite{b14}}   & --    & 63.9          & 62.9         \\ \midrule
\multicolumn{1}{c|}{\multirow{2}{*}{Our systems}} & 2D CNN-BiLSTM          & 1.2~M          & 58.2          & 54.7          \\
\multicolumn{1}{c|}{} & Temporal CNN-BiLSTM   & 200~K & \textbf{63.0}          & 62.0         \\ \bottomrule
\end{tabular}
\end{table}

\subsection{IEMOCAP \& CEMO-4eC$_s$: Emotion detection on 4 classes}

The choice and the performances of our neural architecture Temporal CNN-BiLSTM having been validated on IEMOCAP, we then trained and tested it on the CEMO-4eC$_s$ recordings.
We assessed the speech emotion recognition performance on CEMO-4eC$_s$ (French database) with 4 emotions (Anger, Fear, Positive, Neutral) and as a reference on IEMOCAP (Anglo-American database) with also 4 emotions (Anger, Sadness, Joy, Neutral). We tested the performance of 2 feature sets, with and without the concatenation of  $\Delta\Delta_2$ features and with and without the classification of gender as an auxiliary task (Multitask).

\begin{table}
\centering
\caption{5-fold cross validation scores with Temporal CNN-BiLSTM with or without concatenation of $\Delta\Delta_2$ features and with and without Multitask technique on 4 emotions. For each experiment, the results correspond to the best run for emotion detection but do not correspond to the best gender detection run (which is 94.4\% UA instead of 86.3\%)}
\label{tab:table3}
\begin{tabular}{@{}cc|cccccc@{}}
\toprule
\multicolumn{2}{l}{\multirow{2}{*}{\textbf{Cond. (4 emotions)}}} &
  \multicolumn{3}{|c|}{\textbf{IEMOCAP}} &
  \multicolumn{3}{c}{\textbf{Real-life CEMO-4eC$_s$}} \\
\multicolumn{2}{l}{}                                 & \multicolumn{3}{|c|}{(Eng-US)}           & \multicolumn{3}{c}{(French)} \\ \midrule
\multicolumn{2}{c|}{$\Delta\Delta_2$}                 & -    & -    & \multicolumn{1}{c|}{+}    & -        & -       & +       \\
\multicolumn{2}{c|}{Multitask}                       & -    & +    & \multicolumn{1}{c|}{+}    & -        & +       & +       \\ \midrule
\multicolumn{1}{c|}{\multirow{2}{*}{Gender}} &
  \multicolumn{1}{c|}{\sc{ua} (\%)} &
  -- &
  82.2 &
  \multicolumn{1}{c|}{\textbf{86.3}} &
  -- &
  75.1 &
  \textbf{80.6} \\
\multicolumn{1}{c|}{} & \multicolumn{1}{c|}{\sc{wa} (\%)} & --    & 86.0 & \multicolumn{1}{c|}{87.6} & --        & 79.9    & 85.3    \\ \midrule
\multicolumn{1}{c|}{\multirow{2}{*}{Emotion}} &
  \multicolumn{1}{c|}{\sc{ua} (\%)} &
  61.5 &
  62.3 &
  \multicolumn{1}{c|}{\textbf{63.0}} &
  45.1 &
  44.9 &
  \textbf{45.6} \\
\multicolumn{1}{c|}{} & \multicolumn{1}{c|}{\sc{wa} (\%)} & 61.7 & 61.1 & \multicolumn{1}{c|}{62.0} & 46.1     & 45.2    & 47.1    \\ \bottomrule
\end{tabular}
\end{table}

As can be seen in TABLE~\ref{tab:table3}, the 4 emotion detection task is much more complex on the real-life  database than for the IEMOCAP database. There are also very few differences between performance with Multitask (emotion and gender tasks) compared to the emotion-only task with 4 emotions for both corpora. The concatenation of $\Delta\Delta_2$ parameters slightly improves the system performance, but does not seem very useful with an end-to-end deep learning architecture in our context.

Gender recognition is used here as an auxiliary task to aid SER performance. The gender recognition score 86.3\% (UA) on the spontaneous part of IEMOCAP, is that associated to the best result of SER, which is 63\% (UA). Naturally, our best gender recognition run in the same configuration actually achieves a score of 94.4\% (UA), but with lower SER results.

\subsection{CEMO-4eC$_s$: Emotions detection on 2, 3 and 4 classes}

In an emergency context, the recognition of more than two emotions from call center recordings could be useful for better understanding the situation. Additional tests were performed with the multitask learning technique (emotion and gender) and the $\Delta\Delta_2$ parameters. The temporal CNN-BiLSTM was trained and evaluated respectively on the detection of 4, 3 and 2 emotions in the  CEMO-4eC$_s$ database.

\begin{table}
\centering
\caption{5-fold cross validation scores on Temporal CNN-BiLSTM system for 2, 3 or 4 emotions detection (for each experiment the results correspond to the best run).}
\label{tab:table4}
\begin{tabular}{@{}l|cc@{}}
\toprule
\multicolumn{1}{c|}{\multirow{2}{*}{\textbf{Conditions}}} & \multicolumn{2}{c}{\begin{tabular}[c]{@{}c@{}}\textbf{Real-life CEMO-4eCs}\\ (French)\end{tabular}} \\
\multicolumn{1}{c|}{}                      & \sc{ua} (\%)              & \sc{wa} (\%)              \\ \midrule
{\textbf{4 emotions}:}                          &             &                      \\
Fear, Anger, Positive, Neutral             & \textbf{45.6}                 & 47.1                 \\ \midrule
{\textbf{3 emotions}:}                          &                      &                      \\
Anger, Positive, Neutral                   & 52.4                 & 55.8                 \\
Negative (Anger + Fear), Positive, Neutral & \textbf{54.4}                 & 63.4                 \\ \midrule
{\textbf{2 emotions}:}                          & \multicolumn{1}{l}{} & \multicolumn{1}{l}{} \\
Anger, Neutral                             & 76.9                 & 76.8                 \\
Negative (Anger + Fear), Neutral           & \textbf{77.5}                 & 77.5                 \\
Positive, Negative                         & 69.2                 & 74.4                 \\ \bottomrule
\end{tabular}
\end{table}
The results for the detection of 2 emotions in TABLE~\ref{tab:table4} are above 70\% correct.
It is important to keep in mind that the expressive behaviors of the callers (patient, patient's relatives, or medical staff) could be very different. The detection of 3 and 4 emotions is a significantly more complex task .

\subsection{Within-corpus and cross-corpus emotions recognition (Anger, Neutral)}

When working with realistic emotions, several difficulties appear when trying to make use of multiple corpora or cross-corpus training as the gap between each annotation context may lead to a poor generalization \cite{b22}, \cite{b21}, \cite{b23}.

To perform cross-corpus emotion recognition, we selected the two emotions (Anger and Neutral) common to both the IEMOCAP and CEMO-4eC$_s$ corpora.

\begin{table}
\centering
\caption{5-fold cross validation scores on Temporal CNN-biLSTM system for 2 emotions detection (Anger and Neutral) with IEMOCAP and CEMO-4eC$_s$ using matched training and test conditions. The last entry  assesses the portability of the model trained on IEMOCAP to the CEMO task (i.e crossed conditions using IEMOCAP for training and CEMO-4eC$_s$ for testing). For each experiment we present results corresponding to its best run.}
\label{tab:table5}
\begin{tabular}{@{}cc|c|c@{}}
\toprule
\multicolumn{2}{c|}{\multirow{2}{*}{\textbf{Cond. (Anger vs Neutral)}}} & \textbf{IEMOCAP} & \textbf{Real-life CEMO-4eC$_s$} \\
\multicolumn{2}{c|}{}                                       & (Eng-US) & (French) \\ \midrule
\multicolumn{1}{c|}{\multirow{2}{*}{Matched cond.}}            & \sc{ua} (\%) & 81.1     & 76.9     \\
\multicolumn{1}{c|}{}                             & \sc{wa} (\%) & 79.4     & 76.8     \\ \midrule
\multicolumn{1}{c|}{\multirow{2}{*}{Crossed cond.}} & \sc{ua} (\%) & --    & 61.9     \\
\multicolumn{1}{c|}{}                             & \sc{wa} (\%) &  --    & 61.8     \\ \bottomrule
\end{tabular}
\end{table}

It can be seen in Table~\ref{tab:table5} that the results on the detection of 2 emotions (Anger, Neutral) on both corpora are much closer than was seen for 4 emotions.
The last entry in Table~\ref{tab:table5} assesses the portability of an emotion detection system based on the IEMOCAP data to the real-life CEMO data set. More specifically an experiment was conducted by training the system on IEMOCAP data and using the CEMO-4eC$_s$ data for testing purposes. The results show a notable decline in performance in the cross-corpus experiment; 61.9\% (UA) correct detection of the 2 emotions (Anger, Neutral) was obtained, which is substantially lower than the within-corpus results.  This experiment suggests that the portability of a state-of-the-art system for trained on artificial data is likely to be limited for use in real-life applications, however it is difficult to know how much of the degradation is due to differences in the tasks and languages.

\section{Ethics and replicability}

The use of the CEMO database or any subsets of it carefully respected ethical conventions and agreements ensuring the anonymity of the callers, the privacy of all personal information and the non-diffusion of the audio and meta data including the annotations. The CEMO corpus contains 20 hours of recordings of real conversations between agents and callers obtained following an agreement between an emergency medical center and the LISN-CNRS laboratory \cite{b5}, \cite{b8}.

In order to allow the replicability of our studies, we tested the methods on the widely-used IEMOCAP database and provide here details of the parameters of our experiments.
The classifier choice and its hyperparameters were determined by several tests, primarily based on two SotA research papers \cite{b6} and \cite{b9}. We chose a CNN+BiLSTM system because we needed an neural network architecture capable of processing input spectrogram signals and convolution networks have shown high performance in creating representative features from images. LSTMs are a classical architecture but are effective in detecting long dependencies within a single signal. This strategy is very useful because emotions are often produced in complex ways.
We varied different parameters such as the Fourier transform (Hamming/Hanning window, window size, number of bins per window, window step), we also varied the different parameters of the NN architecture. Specifically for the CNN; the number of convolutions (1 to 5), the kernel size (time dependent or not), the stride and padding. And for the LSTMs; the number of layers, the size of hidden units and the type of outputs of the LSTM (taking either the hidden vector of the last cell or all the output vectors of each cell. We finally concatenated each LSTM representation at each time step because it adds information and helps the dense layers.
All the hyperparameters are listed in Figure 4 for both corpora (IEMOCAP-16kHz/CEMO-8kHz).

All the experiments were carried out using Tensorflow on two GPUs (GeForce GTX 1080 Ti with 11 Gbytes of RAM). We used ReLU activation function \cite{b16} between all layers to benefit from the He Normal Initialization \cite{b15} of our convolution layers. The Adam Optimizer was used with a learning rate schedule based on an Exponential Decay: the initial learning rate is 1e-4, the decay append every 1000 steps an decreased with a rate of 0.9. Our study also used gradient clipping between $-1$ and $1$ to avoid exploding gradients. We choose cross-entropy as the loss function for both tasks. Of course an underlying issue with replicability is the dependence of the results on the amount and order of presentation of the data during the training process and on the conditions of initialization and cross validation procedure. 

\section{Conclusions}
In this work, we illustrate the challenges of the speech emotion recognition task in real-life scenarios such as emergency calls (CEMO-4eC$_s$) through a state-of-the-art NN architecture (Temporal CNN-BiLSTM) first tested on IEMOCAP.
Detecting real-life emotions are clearly more complex than improvised ones, due for example to the large number of speakers (485 for CEMO-4eC$_s$ instead of 10 for IEMOCAP), and lack of ground truth classifications as is reflected by the inter-annotator agreement. The Multitask architecture using emotion and gender and also the use of $\Delta\Delta_2$ in the preprocessing were seen to slightly improve the emotion recognition results. Our system obtained a  63\% (UA) on IEMOCAP with a 5-fold cross validation strategy on 10 speakers and 4 classes. With the same end-to-end deep learning architecture, the performance on the CEMO-4eC$_s$ database are 45.6\% UA for 4 classes (Anger, Fear, Positive, Neutral),  54.4\% UA for 3 classes (Negative, Positive, Neutral) and 77.5\% (UA) for 2  emotions (Negative, Neutral). These results are promising on real-life emotions. 
In conclusion, even if we can reproduce the state-of-the-art of the system on IEMOCAP, the portability of the database is limited for real applications. A similar observation was made on speech recognition where the portability from read to spontaneous speech is widely acknowledged to be limited. The next step will be to propose a multimodal architecture using both the audio signal and the linguistic transcription to improve emotion detection in the context of an emergency call center application.

\section{Acknowledgment}
This PHD thesis is supported by the AI Chair HUMAAINE at LISN-CNRS, led by Laurence Devillers and reuniting researchers in computer science, linguists and behavioral economists from the Paris-Saclay University.


\clearpage
\balance

\end{document}